# Bankruptcy Prediction via Hybrid Resampling and Stacking Ensemble Techniques with Explainable Artificial Intelligence (XAI)-Driven Analysis

**Obu-Amoah Ampomah[1], Edmund Fosu Agyemang[2], Kofi Acheampong[3], Louis Agyekum[4], Enock Adu Bonsu[5], Eric Nyarko[6]**

[1]Department of Statistics, Western Michigan University, Kalamazoo, Michigan, USA
[2]Department of Biostatistics and Data Science, Celia Scott Weatherhead School of Public Health and Tropical Medicine at Tulane University, New Orleans, LA 70112, USA
[3]Department of Economics, Western Michigan University, Kalamazoo, Michigan, USA
[4]Department of Economics, University of Ottawa, Ottawa, ON K1N 6N5, Canada
[5]Department of Epidemiology and Biostatistics, University of Arizona, USA
[6]Department of Statistics and Actuarial Science, University of Ghana, Accra, Ghana
[*]**Correspondence author:** ericnyarko@ug.edu.gh

**Abstract**

This study develops and evaluates a bankruptcy prediction framework that integrates consensus-based feature selection, hybrid resampling, stacking ensembles, and explainable artificial intelligence to improve minority-class detection in severely imbalanced financial data. Using the Taiwanese Bankruptcy Prediction dataset from the UCI Machine Learning Repository, comprising 6,819 firm observations, 95 predictors, and a 3.23% bankruptcy rate, five feature-selection algorithms were first applied, and a consensus retention rule reduced the input space to 23 robust variables. The balanced training data were then generated using SVM-SMOTE, SMOTE-Tomek, and SMOTE-ENN. Five ensemble machine learning classifiers, namely gradient boosting, extreme gradient boosting, histogram-based gradient boosting, LightGBM, and AdaBoost, were compared with five deep learning models, including RNN, LSTM, GRU, DNN, and MLP. In addition, hybrid stacking ensembles combined the five machine learning classifiers as base learners with each deep learning model as a meta-learner. Model performance was assessed using accuracy, recall, specificity, G-mean, and ROC-AUC, while SHAP was used to explain feature contributions. The results show that resampling strategy materially shaped model behavior. SVM-SMOTE and SMOTE-Tomek favored accuracy and specificity, whereas SMOTE-ENN delivered stronger minority-class detection. Among standalone models, the GRU with SMOTE-ENN achieved the best overall predictive balance, with recall of 0.8627, G-mean of 0.8517, and ROC-AUC of 0.9431. Among stacking ensembles, SMOTE-ENN with (GB+XGB+HGB+LGBM+AB) +LSTM provided the strongest compromise between sensitivity and specificity. SHAP analysis identified leverage, profitability, solvency, and operational efficiency indicators as the most influential predictors of bankruptcy risk. These findings support more reliable and interpretable early warning systems for financially distressed firms.

## 1 Introduction

The expanding availability of financial statements and narrative disclosures has intensified the need for accurate bankruptcy predictions for creditors, investors, auditors, and regulators [1, 2]. In applied settings, bankruptcy prediction models estimate firm-level failure risk over future horizons, yet they must operate under severe class imbalance, heterogeneous firm populations, limited observations of actual failure, and instability across industries and macroeconomic conditions. These features make bankruptcy prediction materially different from ordinary static classification, because bankruptcy is a rare event, predictor distributions shift over time, and the evolution of a firm's condition often contains information that one-period cross-sectional models do not capture [3-5]. Accordingly, the recent works treat bankruptcy prediction as a cost-sensitive decision problem in which failing to identify a genuinely distressed firm may be more consequential than incorrectly flagging a healthy one; therefore, model evaluation should emphasize recall, Geometric (G)-mean, area under the curve (AUC), and class-specific error rates rather than raw accuracy alone [6-8]. Recent evidence further shows that predictive performance can be improved by moving beyond traditional ratio-based inputs to incorporate annual-report language, risk-related disclosures, and other unstructured textual signals, while maintaining interpretability for practical decision use [9, 10].

Bankruptcy events are rare relative to non-bankruptcy events, which can lead to trivial but misleading performance if evaluation relies on accuracy alone. For highly imbalanced settings, the precision recall (PR) curve and PR AUC are frequently recommended because they focus on performance on the positive class and can better reflect the practical tradeoff between catching fraud and limiting false alarms [11, 12]. This evaluation perspective has direct methodological consequences. For instance, decision thresholds should be tuned to operational capacity, and training procedures often incorporate class weighting or sampling strategies to increase sensitivity to the minority class without inflating false positives. Synthetic oversampling methods such as Synthetic Minority Over-sampling Technique (SMOTE) are widely used in imbalanced learning to augment minority class examples, although careful validation design is required to avoid leakage in time-dependent fraud streams [13, 14].

Bankruptcy risk signals evolve over time as firms move through changing liquidity, leverage, profitability, and cash flow conditions, while shifts in macroeconomic and industry environments can alter the relationship between predictors and failure outcomes. Consequently, the bankruptcy prediction literature gives considerable attention to temporal stability, dynamic or panel-based modeling, and out-of-time evaluation, since models estimated in one period may generalize poorly when applied to another period with different economic conditions [15, 16]. In practice, bankruptcy risk can be inferred from static accounting and market indicators, multi-period trajectories of financial deterioration, and narrative disclosures in annual reports. Although conventional tabular models remain useful, recent studies show that predictive performance improves when models incorporate textual and other multimodal information, because single-period financial ratios alone provide only a partial view of a firm's evolving condition [9, 10, 17]. This motivates hybrid modeling, where complementary learners target different data modalities and are fused to improve robustness [18, 19].

A host of machine learning (ML) models have been employed for bankruptcy detection. Tree-based ensembles remain a dominant baseline because they capture nonlinear feature interactions, handle mixed variable types, and offer strong performance with limited preprocessing. Gradient boosting machines formalize an additive ensemble framework that incrementally improves predictive performance by fitting weak learners to residual errors [20]. Practical implementations such as XGBoost (Extreme Gradient Boosting) and LightGBM (Light Gradient Boosting Machine) introduced scalable training and engineering optimizations that enable strong performance on large datasets, supporting their widespread adoption in applied fraud analytics [21, 22]. Beyond single-family ensembles, stacked generalization (stacking) combines multiple base learners by training a meta learner on their outputs, aiming to leverage complementary inductive biases. This is particularly relevant for bankruptcy detection because different models can excel under different regimes, such as sparse high-cardinality tabular features versus temporal patterns. Stacking provides a principled framework for hybridization when diverse learners capture different aspects of the fraud signal [23].

Deep learning (DL) has shown promise in bankruptcy prediction when the objective is to learn from the temporal progression of a firm's financial condition rather than from isolated one-period snapshots. Recent evidence indicates that these sequence-aware deep learning approaches can outperform conventional baselines in some settings, especially when multiple periods of firm information are available and the prediction task is explicitly framed in temporal terms [4, 24]. More broadly, recent review evidence identifies deep, hybrid, and time-aware models as an important direction in contemporary bankruptcy prediction research, while also noting that their performance depends on data design, class imbalance treatment, and appropriate temporal validation procedures. This line of work supports a broader conclusion: in many fraud contexts, predictive performance depends on how effectively the model captures short-term deviations from a customer's typical behavior and longer-term patterns of spending. Such dependencies are difficult to represent with single transaction features alone [25, 26].

A hybrid ensemble framework is motivated by the observation that no single modeling family consistently dominates across all fraud conditions. The survey on bankruptcy prediction emphasizes the diversity of challenges and methods, including handling imbalance, drift, and sequential properties, suggesting that integrated solutions may be more robust than single-model approaches [4, 7, 27, 28]. In a hybrid ensemble, model fusion can be implemented through stacking, weighted averaging, or two-stage pipelines where deep models produce representations or risk scores that become inputs to an ensemble meta-learner or otherwise. Stacking is particularly suitable for combining heterogeneous learners because it learns how to weight base models under the validation distribution rather than assuming a fixed combination rule [29]. However, hybridization increases governance complexity. As model pipelines become more layered, institutions face challenges in calibration, monitoring under drift, and communicating rationales to investigators. This directly motivates the integration of explainable artificial intelligence (XAI)-driven analysis as a first-class component of the framework rather than an afterthought.

In bankruptcy detection, model output often triggers consequential actions such as declines, step up authentication, or account restrictions. This makes transparency and interpretability important for auditability, internal governance, and effective human investigation. A prominent perspective argues that relying on black box models with post hoc explanations can create fragile decision pathways in high-stakes contexts, motivating careful design choices regarding interpretability and explanation use [30]. Systematic reviews of explainable AI in finance document rapid growth in XAI adoption across financial tasks, reflecting the need to align predictive models with transparency expectations and stakeholder trust [31]. Two widely adopted post hoc explanation approaches are LIME and SHAP. LIME explains individual predictions by fitting a local surrogate model around a specific instance, providing an interpretable approximation of local behavior [32]. SHAP provides a unified additive feature attribution framework grounded in Shapley values and is frequently used to explain tree ensembles and other complex predictors through consistent feature contribution estimates [33]. In bankruptcy analytics, these tools can support several practical needs: case-level narratives for investigators, global feature importance to validate domain plausibility, and monitoring of explanation stability over time. At the same time, the literature cautions that explanation methods should be assessed for faithfulness and stability, particularly under dataset shift.

Prior studies often emphasize predictive performance on static datasets or focus on a single modeling family, while comparatively fewer works present an end-to-end framework that simultaneously fuses complementary machine learning and deep learning signals [34], evaluates performance using imbalance-appropriate metrics aligned with operational constraints, and produces explanations that remain consistent and actionable as fraud behavior changes over time. In this study, we aim to design and empirically validate a hybrid stacking ensemble framework for bankruptcy detection that combines strong tabular learners (for example, gradient boosting) with deep learning models that capture behavioral dynamics and integrates explainable AI-driven analysis to provide faithful, case-level, and global explanations that support investigation, governance, and monitoring under class imbalance and concept drift.

The remainder of the paper is organized as follows: Section 2 discusses the data and methods used for the study. Section 3 discusses the results and findings of the study, while section 4 concludes the study and provides recommendations for further work.

## 2 Data and Methods

### 2.1 Data Description

The Taiwanese Bankruptcy Prediction dataset is a multivariate business dataset designed for binary classification of firm bankruptcy status. The data were collected from the Taiwan Economic Journal and span the period 1999 to 2009, with bankruptcy defined according to the business regulations of the Taiwan Stock Exchange. The dataset contains 6,819 firm records and 95 predictive features, with the response variable labeled “Bankrupt”. The predictors are primarily financial ratio type variables (for example, profitability, leverage, liquidity, turnover, and growth-related indicators), as reflected in the dataset’s variable list. The UC Irvine (UCI) documentation reports no missing values. A salient characteristic of this dataset is substantial class imbalance: published analyses report approximately 3.226% bankrupt and 96.774% non-bankrupt firms, indicating that evaluation should emphasize imbalance-aware metrics (for example, Receiver Operating Characteristic curve (ROC)-AUC, recall, and F-measures) rather than accuracy alone. The dataset is distributed via the UCI Machine Learning Repository with DOI 10.24432/C5004D under a CC BY 4.0 license as provided in and may be assessed via

https://archive.ics.uci.edu/dataset/572/taiwanese+bankruptcy+prediction. No missing values were observed in the dataset.

Table 1 presents an overview of recent studies which employed ML and DL models for bankruptcy prediction tasks.

Table 1: Overview of recent studies on bankruptcy prediction using ML and DL models

| Author, Year | Methods | Number of Variables | Sample Size | Interpretable Methods | Major Findings |
|---|---|---|---|---|---|
| Du Jardin, [35], 2021 | Biclustering, Neural Network-based Ensembles | 44 | 700 bankrupt, 34,300 non-bankrupt firms | None | The ensemble approach that integrates biclustering with neural networks enhanced prediction accuracy. |
| Kim et al., [28], 2022 | RNN, LSTM and Ensemble | 8 | 454,752 firm-month observations, 2057 of which are a firm's bankruptcy | None | Compared with traditional techniques, RNN and LSTM methods showed superior performance in handling sequential financial data. |
| Jabeur & Serret, [36], 2023 | Fuzzy set qualitative comparative analysis (fsQCA) and convolutional neural networks (CNN) | 17 | 133 bankrupt, 133 non-bankrupt firms | None | Fuzzy CNN led to better performance than when using traditional methods. |
| Mattos et al., [37], 2024 | Logistic Regression, SVM, AdaBoost, XGBoost, Bagging, and Random Forest | 26 | 503 private firms | Tree-based feature importance; permutation importance | Machine learning models outperformed logistic regression in predicting reorganization outcomes. Financial ratios were less informative under low-quality accounting information, while institutional factors and financial-information-quality proxies were more influential. |
| Nguyen et al., [38], 2025 | Logistic Regression, Discriminant Analysis, SVM, Neural Network, Random Forest, LightGBM, XGBoost, and NGBoost | 24 | 4,054 bankrupt firms and 5,600 non-failed firms | SHAP | XGBoost achieved the best predictive performance across the 1-to-5-year horizons, with ensemble models generally outperforming traditional models. SHAP identified the tax ratio, interest level, debtor days as among the most influential predictors of bankruptcy. |
| Lin et al., [39], 2025 | LASSO feature selection, CNN | 35 | 2,915 bankrupt and 15,121 non-bankrupt firms | SHAP; LIME | The CNN model based on grayscale financial ratio images outperformed traditional bankruptcy prediction methods, while SHAP and LIME improved interpretability by identifying influential financial indicators. |

### 2.2 Data Preprocessing and method of analysis

Data preprocessing is a critical stage in ML workflows. To ensure data quality, duplicate observations were identified and removed. Preprocessing comprised the encoding of categorical target, the scaling of continuous variables, and the application of class-balancing procedures. The target variable, Bankrupt, was transformed into a numerical format using *LabelEncoder* from *sklearn.preprocessing*. The predictor variables were normalized using *MinMaxScaler* to rescale the feature values into the range of 0 to 1. The *scaler* was fitted on the training data and subsequently applied to both the training and test sets to ensure that all variables were measured on a comparable scale. This preprocessing step is particularly important for ML and DL algorithms that are sensitive to differences in feature magnitude, as it helps improve numerical stability and supports more reliable model training and evaluation.

Five feature selection techniques presented in Table 1 were applied to identify the most informative predictors for subsequent modelling. Specifically, the study employed metaheuristic, nature-inspired optimization algorithms that iteratively explore the feature space and converge toward near-optimal subsets under a predefined objective function (for example, maximizing predictive performance while minimizing the number of selected variables). Feature selection is essential because removing irrelevant or redundant features can enhance classifier generalization, reduce overfitting, improve computational efficiency, and increase interpretability [40]. Randomized search cross validation was employed to identify the optimal hyperparameter combinations because it offers an efficient and robust alternative to exhaustive grid search, particularly when the search space is large and computationally demanding. This approach enables a broader exploration of candidate parameter settings within a limited computational budget, while cross validation helps ensure that the selected configuration generalizes well and reduces the risk of overfitting [41]. These optimal hyperparameters were used in training the ML models employed in the study.

The bankruptcy dataset was partitioned into training and testing subsets using an 80:20 ratio. A severe class imbalance was observed, which may introduce bias in the predictions because the models can become predisposed toward the majority class [13, 14]. To address this issue, three hybrid resampling techniques were adopted: SVM-SMOTE [42], SMOTE-Tomek [14] and SMOTE-Edited Nearest Neighbors (ENN)[43]. In this study, five (5) ensemble learning models: gradient boosting (GB), extreme gradient boosting (XGB), Histogram-based gradient boosting (HGB), light extreme gradient machines (LGBM), and adaptive boosting (AB) were employed as base learners. Additionally, five (5) DL models, including recurrent neural network (RNN), long short-term memory (LSTM), gated recurrent units (GRU), deep neural networks (DNN), and multi-layer perceptron (MLP), were employed as meta-learners.

To improve model interpretability, we applied a well-known XAI technique: SHAP to explain the contribution of each feature to the model's predictions. SHAP computes feature attributions using Shapley values, allowing both global and local interpretation of model behavior. This enabled the identification of the most influential financial indicators and provided insight into how specific features increased or decreased the likelihood of bankruptcy classification [44, 45].

### 2.3 Model Performance Evaluation Metrics

All ten (10) models (ML and DL) were evaluated mainly using the test accuracy, recall, specificity, area under the receiver operating curve (ROC-AUC) and G-mean.

The computational formulas are given respectively by (1)-(5):

$$\text{Accuracy} = \frac{\text{TP+TN}}{\text{TP+TN + FP+FN}} \tag{1}$$

$$\text{Recall (Sensitivity)} = \frac{\text{TP}}{\text{TP + FN}} \tag{2}$$

$$\text{Specificity} = \frac{\text{TN}}{\text{TN + FP}} \tag{3}$$

$$\text{ROC-AUC} = \int_0^1 \text{TPR(FPR)}\ d(\text{FPR}) \tag{4}$$

$$\text{G-mean}=(\text{Recall} \times \text{Specificity})^{1/2} \tag{5}$$

The empirical workflow adopted for the study is given in Figure 1

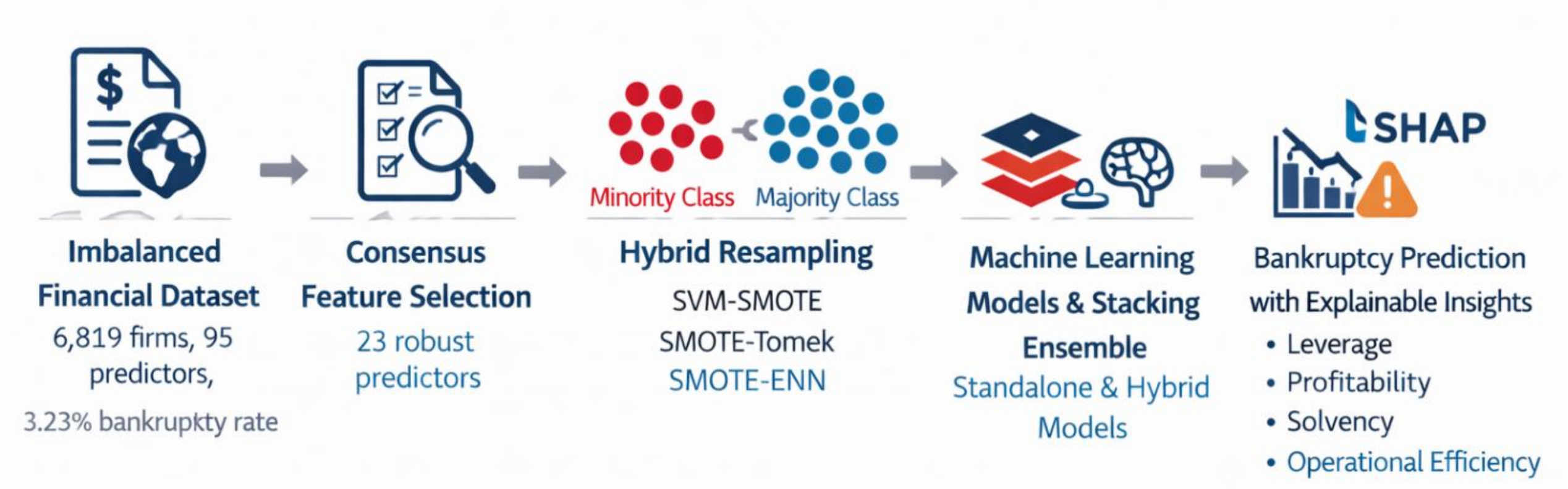

**Figure 1: Empirical workflow adopted for the study**

Algorithm 1 presents the formulation of hybrid-SMOTE-based model training.

**Algorithm 1: Hybrid-SMOTE-Based Boosting Model Training**

**Input:** X_train_scaled, X_test_scaled, y_train, y_test
Apply **Hybrid-SMOTE** to (X_train_scaled, y_train)
    X_train_resampled, y_train_resampled <-fit_resample(X_train_scaled, y_train)
Define candidate models and hyperparameter search spaces for GB, XGB, HGB, LGBM, and AB
For each model m in the candidate set **do**
    Build a pipeline with classifier m
    Tune m using RandomizedSearchCV (n_iter = 5, cv = 5, scoring = ROC-AUC, refit = True)
    Fit the tuned model on (X_train_resampled, y_train_resampled)
    Predict class labels and scores on X_test_scaled
    Compute Accuracy, Precision, Recall, Specificity, F1-score, G-mean, and ROC-AUC
**End for**
Rank models by ROC-AUC, G-mean, F1-score, and Accuracy
Return the best models and the summary results table

**NB:** Hybrid SMOTE was alternated using SVM-SMOTE, SMOTE-Tomek and SMOTE-ENN

### 2.4 Feature Selection

Due to the large number of features (95), feature selection techniques were employed to select the best features for the analysis. Table 2 lists the number of features selected by the 5 different feature selection algorithms.

**Table 2: Feature Selection Results using the 5 feature selection algorithms**

| Feature selection method used | Number of features chosen | Features chosen | Validation ROC-AUC |
|---|---|---|---|
| Whale Optimization (WOA) [46] | 7 | ROA(C) before interest and depreciation before interest, Operating Gross Margin, Persistent EPS in the Last Four Seasons, Quick Ratio, Net worth/Assets, Retained Earnings to Total Assets, Current Asset Turnover Rate | 0.9112 |
| Particle Swarm Optimization (PSO) [47] | 20 | ROA(C) before interest and depreciation before interest, Operating Gross Margin, Operating Expense Rate, Interest-bearing debt interest rate, Persistent EPS in the Last Four Seasons, Operating Profit Growth Rate, Total debt/Total net worth, Net worth/Assets, Operating profit/Paid-in capital, Net profit before tax/Paid-in capital, Average Collection Days, Retained Earnings to Total Assets, Total expense/Assets, Current Asset Turnover Rate, Cash Turnover Rate, Cash Flow to Total Assets, Net Income to Total Assets, Total assets to GNP price, Gross Profit to Sales, Equity to Liability | 0.9140 |
| Salp Swarm Optimization (SSO) [48] | 20 | ROA(C) before interest and depreciation before interest, Realized Sales Gross Margin, Operating Expense Rate, Per Share Net profit before tax (Yuan ¥), Operating Profit Growth Rate, Current Ratio, Debt ratio %, Net worth/Assets, Operating profit/Paid-in capital, Net profit before tax/Paid-in capital, Total Asset Turnover, Inventory/Working Capital, Retained Earnings to Total Assets, Total expense/Assets, Net Income to Total Assets, Total assets to GNP price, Gross Profit to Sales, Net Income to Stockholder's Equity, Interest Coverage Ratio (Interest expense to EBIT), Equity to Liability | 0.9004 |
| Bat Algorithm (BA) [49] | 20 | ROA(C) before interest and depreciation before interest, Operating Gross Margin, Net Value Per Share (A), Persistent EPS in the Last Four Seasons, Operating Profit Growth Rate, Net worth/Assets, Operating profit/Paid-in capital, Net profit before tax/Paid-in capital, Total Asset Turnover, Accounts Receivable Turnover, Average Collection Days, Current Liability to Assets, Inventory/Working Capital, Retained Earnings to Total Assets, Total expense/Assets, Current Asset Turnover Rate, Cash Flow to Total Assets, Net Income to Total Assets, Gross Profit to Sales, Net Income Flag | 0.9022 |
| Mutual Information (MI) [50] | 20 | ROA(C) before interest and depreciation before interest, ROA(A) before interest and % after tax, Pre-tax net Interest Rate, Continuous interest rate (after tax), Net Value Per Share (B), Persistent EPS in the Last Four Seasons, Per Share Net profit before tax (Yuan ¥), Interest Expense Ratio, Debt ratio %, Net worth/Assets, Borrowing dependency, Net profit before tax/Paid-in capital, Retained Earnings to Total Assets, Total income/Total expense, Net Income to Total Assets, Net Income to Stockholder's Equity, Liability to Equity, Degree of Financial Leverage (DFL), Interest Coverage Ratio (Interest expense to EBIT), Equity to Liability | 0.5417 |

### 2.4.1 Consensus Based Feature Retention Criterion

To promote a predictor set that is both robust and not tied to any single algorithm, a consensus criterion was applied across the five feature selection approaches (Figure 2). The premise is that variables identified repeatedly by multiple, conceptually distinct methods are less likely to reflect idiosyncrasies of a particular scoring rule or search strategy and more likely to capture a stable underlying signal. Accordingly, only predictors exhibiting substantial cross method agreement were retained for subsequent modelling and substantive interpretation

A predictor was retained only when it was identified by at least two of the five feature selection algorithms. In doing so, the procedure supports both parsimony and robustness by reducing the feature space while emphasizing predictors that remain consistently informative across diverse selection frameworks.

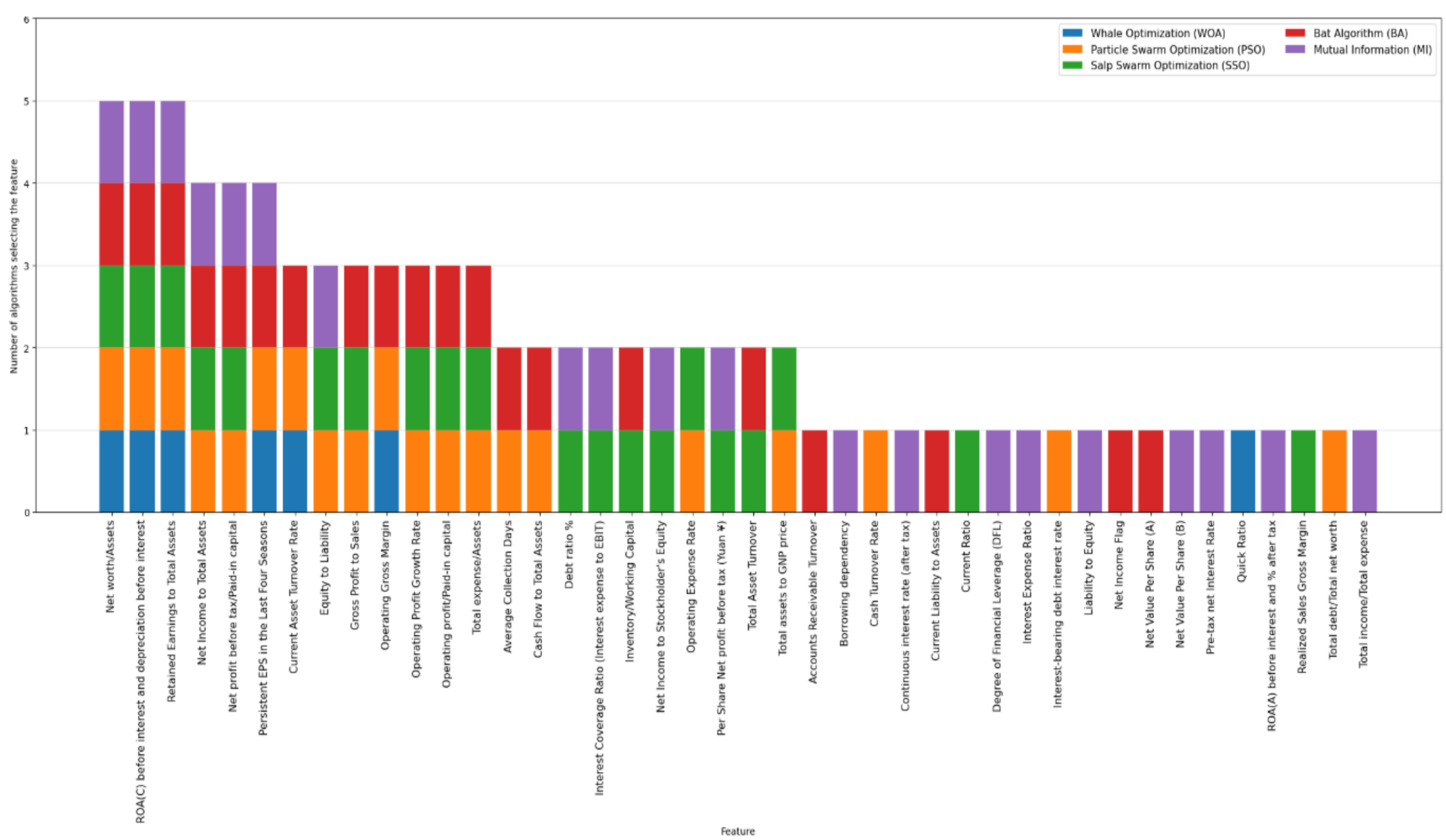

**Figure 2: Feature selection frequency across algorithms (stacked contributions)**

From Figure 2, using the consensus-based feature retention criterion, 23 features (frequency ≥ 2) namely Net worth/Assets; ROA(C) before interest and depreciation before interest; Retained Earnings to Total Assets; Net Income to Total Assets; Net profit before tax/Paid-in capital; Persistent EPS in the Last Four Seasons; Current Asset Turnover Rate; Equity to Liability; Gross Profit to Sales; Operating Gross Margin; Operating Profit Growth Rate; Operating profit/Paid-in capital; Total expense/Assets; Average Collection Days; Cash Flow to Total Assets; Debt ratio %; Interest Coverage Ratio (Interest expense to EBIT); Inventory/Working Capital; Net Income to Stockholder's Equity; Operating Expense Rate; Per Share Net profit before tax (Yuan ¥); Total Asset Turnover and Total assets to GNP price were retained for the analysis.

## 3 Results and Findings

Figure 3 presents the bankruptcy class distribution before and after the application of hybrid resampling techniques. The original dataset was highly imbalanced, with 6,599 non-bankruptcy cases (96.77%) and only 220 bankruptcy cases (3.23%), as shown in Figure 3(a). Under such conditions, pure undersampling would likely be unsuitable because it would discard a large portion of the majority class, potentially resulting in information loss and reduced generalizability. To address this issue, hybrid oversampling methods were applied to improve minority-class representation while retaining as much useful information as possible. Figures 3(b) and 3(c) show that SVM-SMOTE and SMOTE-Tomek achieved balanced class distributions, with 5,286 and 5,279 instances per class, respectively. By contrast, SMOTE-ENN produced a slightly uneven distribution of 4,662 non-bankruptcy and 5,234 bankruptcy cases (as shown in Figure 3(d)), reflecting the effect of ENN in removing noisy or misclassified samples after resampling. These findings indicate that the adopted hybrid resampling methods effectively mitigated the severe class imbalance and produced training distributions that are more suitable for robust bankruptcy prediction.

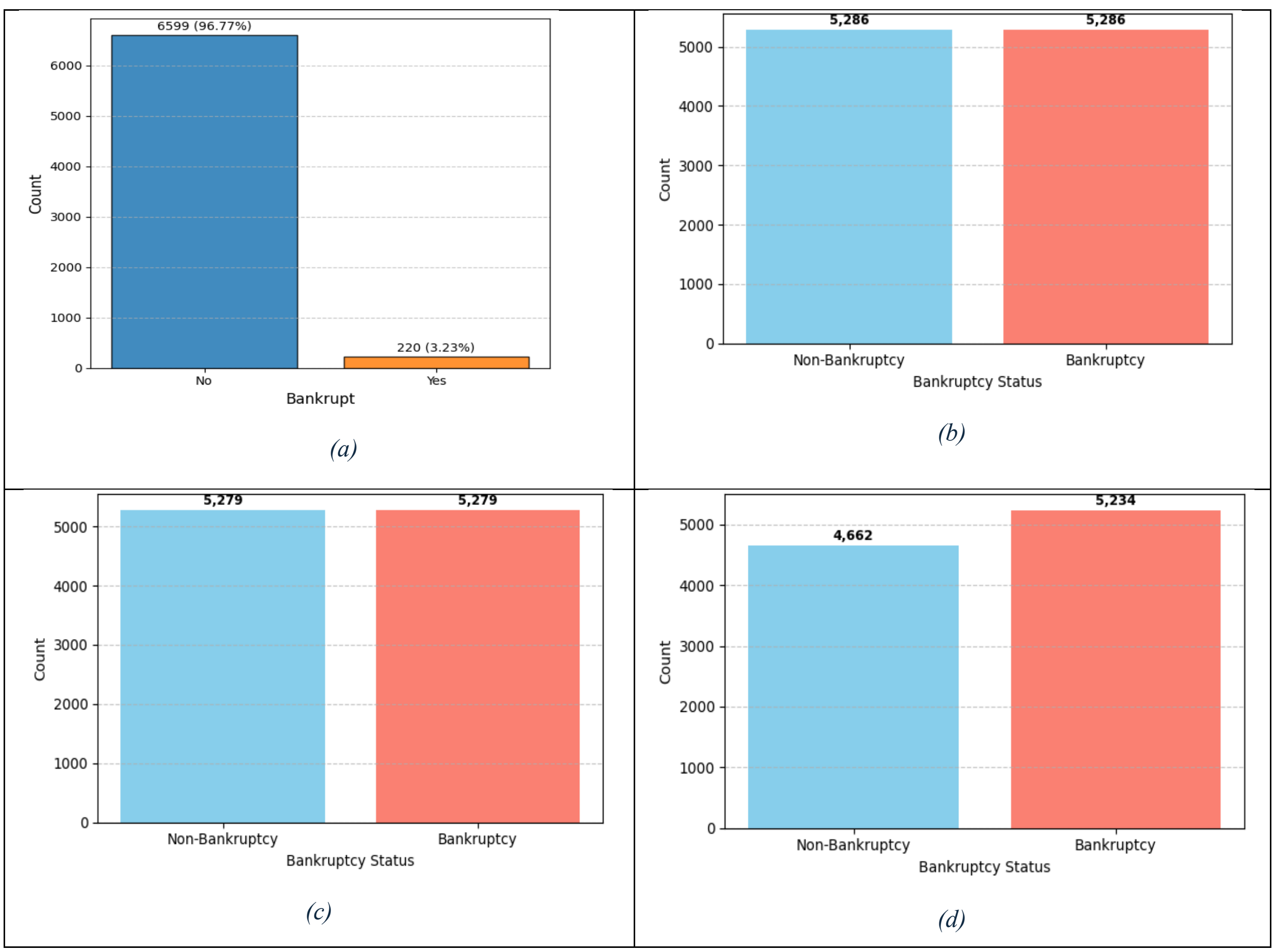

**Figure 3: Bar charts showing bankruptcy data counts before and after application of hybrid resampling techniques. (a) Bankruptcy class distribution of original imbalanced dataset (b) Bankruptcy class distribution after SVM-SMOTE (c) Bankruptcy class distribution after SMOTE–Tomek (d) Bankruptcy class distribution after SMOTE–ENN**

In Table 3, the optimized hyperparameter settings were derived from a targeted search informed by previous studies, best practices in model development, and the need to balance predictive accuracy with generalization. Across the boosting-based classifiers, the tuning process focused primarily on parameters that influence learning pace, ensemble strength, and tree complexity. The number of estimators and learning rate were varied to identify combinations that allowed the models to learn effectively without becoming overly sensitive to the training data. Similarly, depth- and split-related parameters were adjusted to reduce the likelihood of overfitting while preserving the ability to capture meaningful patterns in the data. The results demonstrate that all tuned ensemble models performed strongly, although Extreme Gradient Boosting produced the best overall discrimination performance, with a CV ROC-AUC of 0.9349. This was followed closely by Histogram-based Gradient Boosting (0.9336) and Gradient Boosting (0.9324), which also showed competitive performance. These findings indicate that ensemble boosting approaches were particularly well suited to the classification task addressed in this study. No hyperparameter tuning was done for the DL models.

**Table 3: Optimized configuration and optimal value of hyperparameters for all compared ML models**

| Models | Hyperparameter | Testing Range | Optimal Value | Best CV ROC-AUC |
|---|---|---|---|---|
| **Gradient Boosting** | n_estimators | 100, 200, 400 | 200 | 0.9324 |
| | learning_rate | 0.01, 0.05, 0.1 | 0.1 | |
| | max_depth | 2, 3, 4 | 4 | |
| | min_samples_split | 2, 5, 10 | 10 | |
| | min_samples_leaf | 1, 2, 4 | 2 | |
| **Extreme Gradient Boosting** | n_estimators | 200, 400, 800 | 400 | 0.9349 |
| | learning_rate | 0.01, 0.05, 0.1 | 0.1 | |
| | max_depth | 3, 5, 7 | 5 | |
| | min_child_weight | 1, 5, 10 | 10 | |
| | colsample_bytree | 0.6, 0.8, 1.0 | 1.0 | |
| **Histogram-based Gradient Boosting** | learning_rate | 0.01, 0.05, 0.1 | 0.1 | 0.9336 |
| | max_depth | 2, 3, 5 | 2 | |
| | l2_regularization | 0.0, 0.1, 1.0 | 1.0 | |
| | early_stopping | True | True | |
| | validation_fraction | 0.1, 0.2 | 0.1 | |
| **Light Extreme Gradient Machines** | n_estimators | 200, 400, 800 | 200 | 0.9323 |
| | learning_rate | 0.01, 0.05, 0.1 | 0.05 | |
| | max_depth | -1, 3, 5, 7 | -1 | |
| | min_child_samples | 10, 20, 50 | 10 | |
| | subsample | 0.7, 0.85, 1.0 | 0.7 | |
| **Adaptive Boosting** | n_estimators | 50, 100, 200, 400 | 400 | 0.9262 |
| | learning_rate | 0.01, 0.1, 1.0 | 1.0 | |
| | algorithm | SAMME, SAMME.R | SAMME | |

In Table 4, the SVM-SMOTE resampling technique produced distinct performance differences between machine learning and deep learning models. The machine learning models generally achieved higher accuracy and specificity, with LGBM recording the best accuracy (0.9545) and specificity (0.9726), indicating stronger classification of non-bankrupt firms. However, the deep learning models showed better sensitivity to the minority class. MLP achieved the highest recall (0.8235) and G-mean (0.8574), while HGB recorded the highest ROC-AUC (0.9408) and highest recall (0.7451) among the ML models. These results indicate that, although some ML models were more effective in overall classification, the DL models provided better balance in detecting bankrupt firms, which is more important in imbalanced bankruptcy prediction.

**Table 4: Performance evaluation metrics of ML and DL models for bankruptcy prediction with SVM-SMOTE resampling technique**

| Models | Accuracy | Recall | Specificity | G-mean | ROC-AUC |
|---|---|---|---|---|---|
| GB | 0.9428 | 0.5882 | 0.9566 | 0.7501 | 0.9128 |
| XGB | 0.9538 | 0.5882 | 0.9680 | 0.7546 | 0.8964 |
| HGB | 0.9172 | 0.7451 | 0.9238 | 0.8297 | 0.9408 |
| LGBM | 0.9545 | 0.4902 | 0.9726 | 0.6905 | 0.9170 |
| AB | 0.9326 | 0.7255 | 0.9406 | 0.8261 | 0.9348 |
| RNN | 0.8996 | 0.7843 | 0.9040 | 0.8421 | 0.9377 |
| LSTM | 0.9172 | 0.6863 | 0.9261 | 0.7972 | 0.9121 |
| GRU | 0.8930 | 0.7647 | 0.8979 | 0.8287 | 0.9224 |
| DNN | 0.9106 | 0.7059 | 0.9185 | 0.8052 | 0.9333 |
| MLP | 0.8900 | 0.8235 | 0.8926 | 0.8574 | 0.9400 |

Figure 4 presents the resulting ROC and AUC of ML and DL models after resampling with SVM-SMOTE technique.

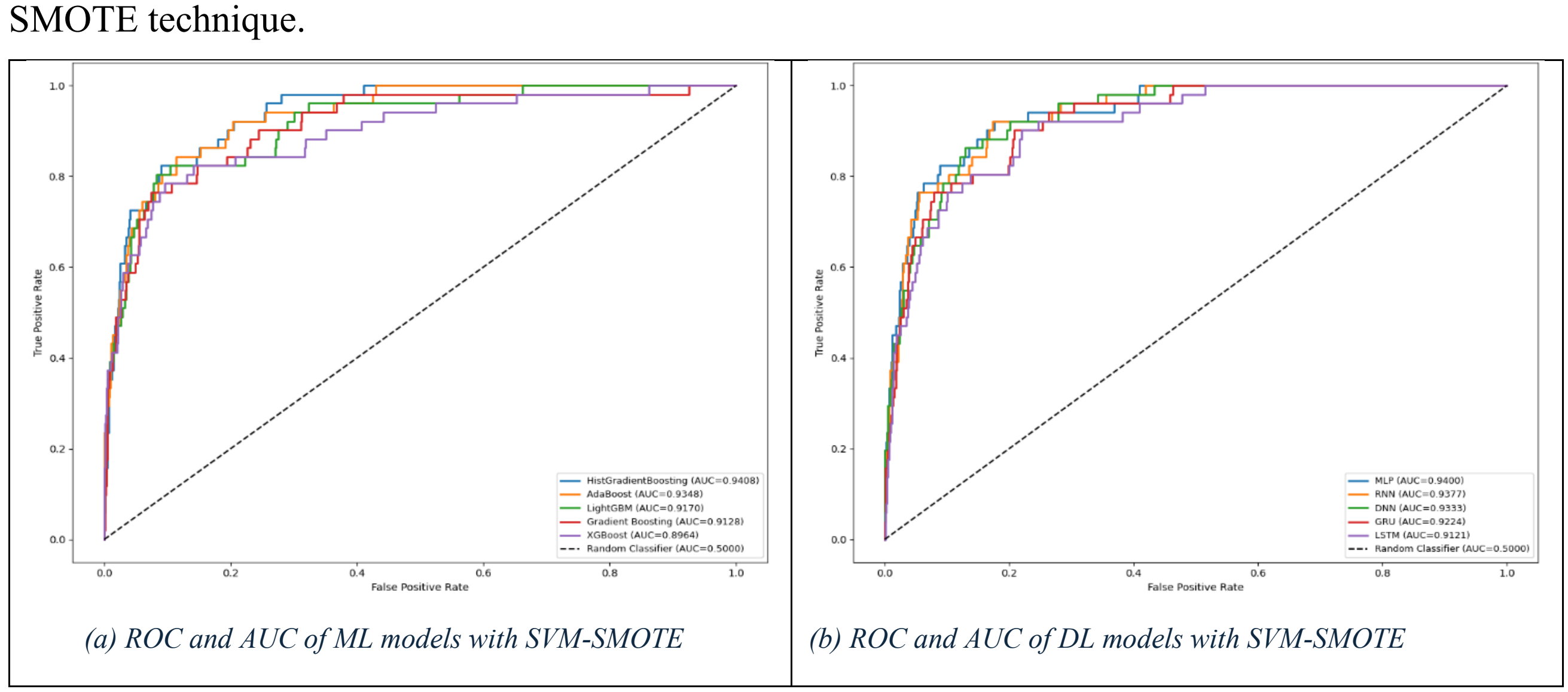

*(a) ROC and AUC of ML models with SVM-SMOTE*

*(b) ROC and AUC of DL models with SVM-SMOTE*

**Figure 4: ROC and AUC of ML and DL models for bankruptcy prediction with SVM-SMOTE resampling technique.**

The results reported in Table 5 show that the SMOTE-Tomek resampling method produced noticeable differences in the predictive behavior of the ML and DL models used for bankruptcy prediction. The ML models generally achieved higher accuracy and specificity, with LGBM producing the best accuracy (0.9494) and specificity (0.9673), indicating strong performance in identifying non-bankrupt firms. HGB recorded the highest ROC-AUC (0.9274) and highest recall (0.7843) among the ML models. However, the DL models showed stronger recall, G-mean, and ROC-AUC, suggesting better capability in detecting bankrupt firms and maintaining balanced predictive performance. Notably, MLP achieved the highest recall (0.8627) and ROC-AUC (0.9327), while RNN recorded the highest G-mean (0.8481). These findings imply that, although the ML models were more effective in overall classification accuracy, the DL models were more suitable for bankruptcy prediction under SMOTE-Tomek because they better captured the minority bankrupt class, which is more critical in imbalanced financial datasets.

**Table 5: Performance evaluation metrics of ML and DL models for bankruptcy prediction with SMOTE-Tomek resampling technique**

| Models | Accuracy | Recall | Specificity | G-mean | ROC-AUC |
|---|---|---|---|---|---|
| GB | 0.9289 | 0.6275 | 0.9406 | 0.7682 | 0.9044 |
| XGB | 0.9472 | 0.5294 | 0.9634 | 0.7142 | 0.9055 |
| HGB | 0.8871 | 0.7843 | 0.8911 | 0.8360 | 0.9274 |
| LGBM | 0.9494 | 0.4902 | 0.9673 | 0.6886 | 0.9177 |
| AB | 0.8952 | 0.7451 | 0.9010 | 0.8193 | 0.9122 |
| RNN | 0.9120 | 0.7843 | 0.9170 | 0.8481 | 0.9306 |
| LSTM | 0.8746 | 0.7843 | 0.8781 | 0.8299 | 0.9163 |
| GRU | 0.8592 | 0.8039 | 0.8614 | 0.8322 | 0.9236 |
| DNN | 0.8490 | 0.8235 | 0.8500 | 0.8366 | 0.9193 |
| MLP | 0.8372 | 0.8627 | 0.8363 | 0.8494 | 0.9327 |

Figure 5 presents the resulting ROC and AUC of ML and DL models after resampling with SMOTE-Tomek technique.

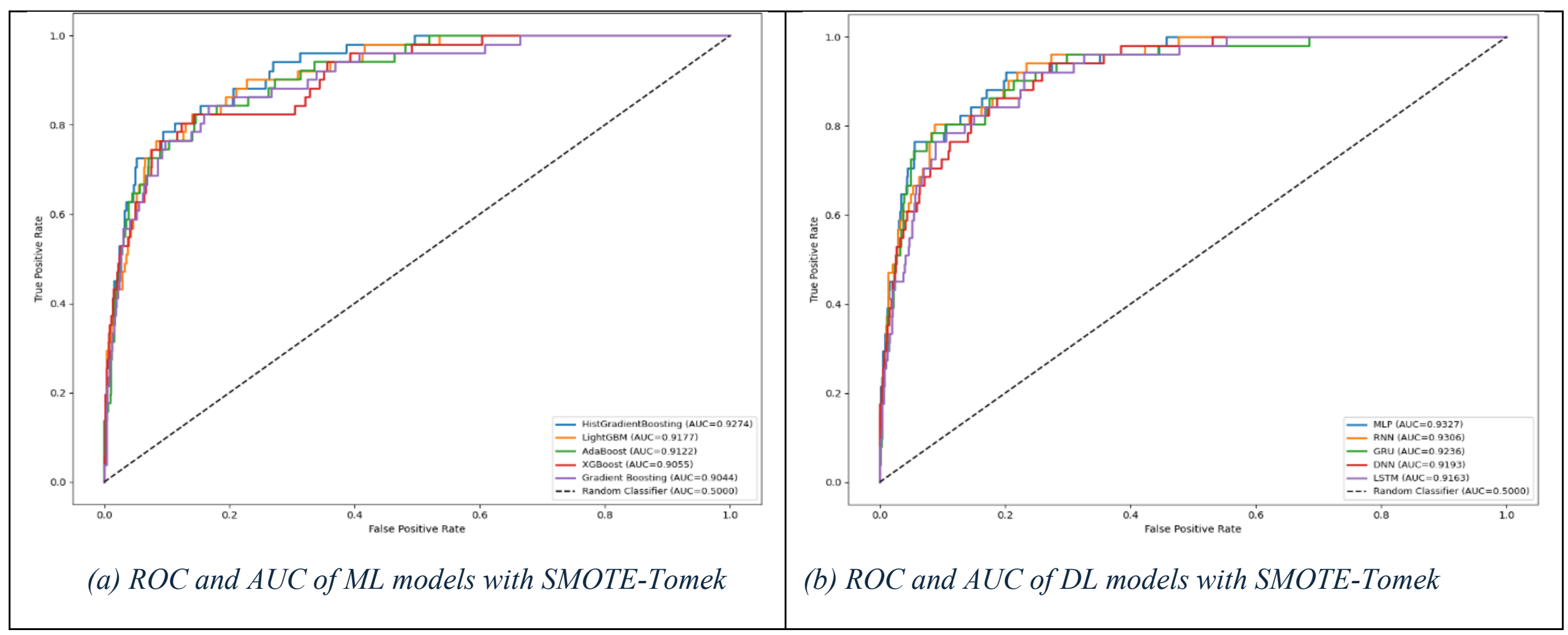

*(a) ROC and AUC of ML models with SMOTE-Tomek*

*(b) ROC and AUC of DL models with SMOTE-Tomek*

**Figure 5: ROC and AUC of ML and DL models for bankruptcy prediction with SMOTE-Tomek resampling technique.**

The findings in Table 6 show that the SMOTE-ENN resampling technique affected ML and DL models differently in bankruptcy prediction. The ML models generally achieved higher accuracy and specificity, with LGBM producing the best overall accuracy (0.9289) and specificity (0.9383), indicating strong performance in identifying non-bankrupt firms. HGB again recorded the highest ROC-AUC (0.9260) and highest recall (0.8039) among the ML models. However, the DL models outperformed the ML models in recall, G-mean, and ROC-AUC, suggesting greater effectiveness in detecting bankrupt firms and maintaining balanced predictive performance. Notably, LSTM and MLP achieved the highest recall (0.8824), RNN produced the highest G-mean (0.8575), and GRU recorded the best ROC-AUC (0.9431). These results suggest that, although ML models were stronger in overall classification accuracy, DL models were more suitable for bankruptcy prediction under SMOTE-ENN because they provided better minority-class detection, which is more critical in imbalanced financial distress datasets.

**Table 6: Performance evaluation metrics of ML and DL models for bankruptcy prediction with SMOTE-ENN resampling technique**

| Models | Accuracy | Recall | Specificity | G-mean | ROC-AUC |
|---|---|---|---|---|---|
| GB | 0.9091 | 0.7451 | 0.9155 | 0.8259 | 0.9034 |
| XGB | 0.9164 | 0.7255 | 0.9238 | 0.8187 | 0.9039 |
| HGB | 0.8732 | 0.8039 | 0.8759 | 0.8391 | 0.9260 |
| LGBM | 0.9289 | 0.6863 | 0.9383 | 0.8025 | 0.9145 |
| AB | 0.8761 | 0.7647 | 0.8804 | 0.8205 | 0.9095 |
| RNN | 0.8526 | 0.8627 | 0.8522 | 0.8575 | 0.9394 |
| LSTM | 0.8006 | 0.8824 | 0.7974 | 0.8388 | 0.9212 |
| GRU | 0.8416 | 0.8627 | 0.8408 | 0.8517 | 0.9431 |
| DNN | 0.8438 | 0.8431 | 0.8439 | 0.8435 | 0.9352 |
| MLP | 0.8277 | 0.8824 | 0.8256 | 0.8525 | 0.9390 |

Figure 6 presents the resulting ROC and AUC of ML and DL models after resampling with SMOTE-ENN technique.

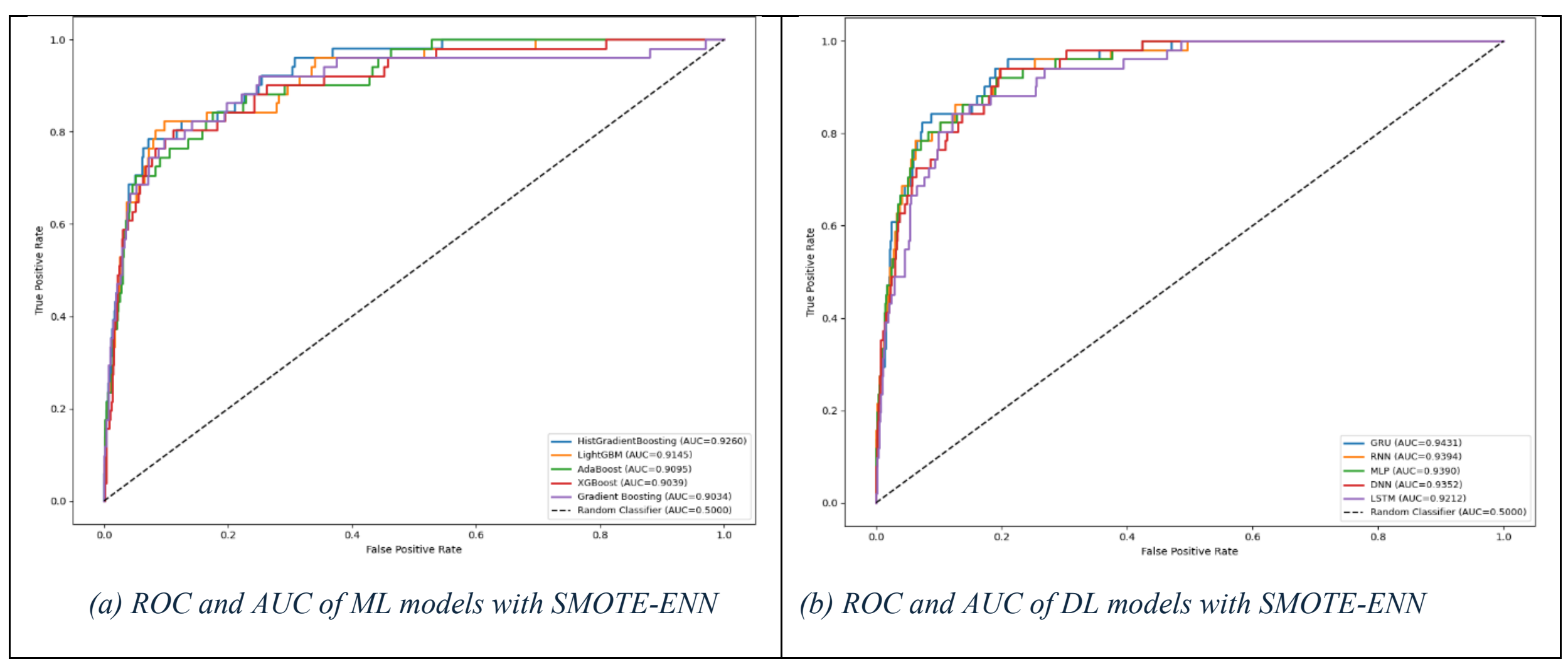

*(a) ROC and AUC of ML models with SMOTE-ENN*

*(b) ROC and AUC of DL models with SMOTE-ENN*

**Figure 6: ROC and AUC of ML and DL models for bankruptcy prediction with SMOTE-ENN resampling technique.**

A descriptive statistical analysis of **Table 7** shows that the hybrid stacking ensembles exhibit a clear trade-off between overall classification accuracy and minority-class detection, depending on the resampling strategy used. Across the three resampling methods, the SMOTE-SVM variants produced the strongest average accuracy (mean = 0.9576) and specificity (mean = 0.9785), followed closely by SMOTE-Tomek (accuracy mean = 0.9559; specificity mean = 0.9758). In contrast, SMOTE-ENN yielded the lowest average accuracy (mean = 0.9361) and specificity (mean = 0.9461), but it achieved the best recall (mean = 0.6784) and the highest average G-mean (mean = 0.8006). This pattern indicates that SMOTE-ENN substantially improved the detection of bankrupt firms, although this gain came at the cost of more false positives among non-bankrupt firms.

The results in Table 6 reveal that model performance varied systematically across resampling strategies. SMOTE-SVM produced the highest average accuracy and specificity, indicating superior majority-class classification, whereas SMOTE-ENN achieved the highest recall and G-mean, reflecting better minority-class detection and more balanced predictive performance. Among all hybrid stacking ensembles, SMOTE-ENN+(GB+XGB+HGB+LGBM+AB)+LSTM emerged as the most practically effective model, achieving recall of 0.7255, G-mean of 0.8254, and ROC-AUC of 0.9270. These findings suggest that, although SMOTE-SVM favors overall correctness, SMOTE-ENN is more appropriate for bankruptcy prediction because it improves the identification of financially distressed firms, which is the more critical classification objective in imbalanced datasets.

**Table 7: Performance metrics of hybrid stacking ensemble models (5+1) bankruptcy prediction with SMOTE-SVM, SMOTE-Tomek, and SMOTE-ENN resampling techniques with ML models as base learners and DL models as meta learners**

| Models | Accuracy | Recall | Specificity | G-mean | ROC-AUC |
|---|---|---|---|---|---|
| **Testing Performance-SMOTE-SVM (5+1)** | | | | | |
| (GB+XGB+HGB+LGBM+AB)+RNN | 0.9619 | 0.3529 | 0.9855 | 0.5898 | 0.8841 |
| (GB+XGB+HGB+LGBM+AB)+LSTM | 0.9523 | 0.4706 | 0.9711 | 0.6760 | 0.9324 |
| (GB+XGB+HGB+LGBM+AB)+GRU | 0.9538 | 0.4902 | 0.9718 | 0.6902 | 0.9325 |
| (GB+XGB+HGB+LGBM+AB)+DNN | 0.9626 | 0.3529 | 0.9863 | 0.5900 | 0.9264 |
| (GB+XGB+HGB+LGBM+AB)+MLP | 0.9575 | 0.4314 | 0.9779 | 0.6495 | 0.7371 |
| **Testing Performance-SMOTE-Tomek (5+1)** | | | | | |
| (GB+XGB+HGB+LGBM+AB)+RNN | 0.9604 | 0.3333 | 0.9848 | 0.5729 | 0.6599 |
| (GB+XGB+HGB+LGBM+AB)+LSTM | 0.9457 | 0.5490 | 0.9612 | 0.7264 | 0.9256 |
| (GB+XGB+HGB+LGBM+AB)+GRU | 0.9604 | 0.4510 | 0.9802 | 0.8173 | 0.6649 |
| (GB+XGB+HGB+LGBM+AB)+DNN | 0.9575 | 0.3922 | 0.9794 | 0.6198 | 0.7671 |
| (GB+XGB+HGB+LGBM+AB)+MLP | 0.9553 | 0.4902 | 0.9733 | 0.6907 | 0.6650 |
| **Testing Performance-SMOTE-ENN (5+1)** | | | | | |
| (GB+XGB+HGB+LGBM+AB)+RNN | 0.9450 | 0.6078 | 0.9581 | 0.7631 | 0.7903 |
| (GB+XGB+HGB+LGBM+AB)+LSTM | 0.9311 | 0.7255 | 0.9391 | 0.8254 | 0.9270 |
| (GB+XGB+HGB+LGBM+AB)+GRU | 0.9304 | 0.7255 | 0.9383 | 0.8251 | 0.9268 |
| (GB+XGB+HGB+LGBM+AB)+DNN | 0.9413 | 0.6667 | 0.9520 | 0.7967 | 0.8145 |
| (GB+XGB+HGB+LGBM+AB)+MLP | 0.9326 | 0.6667 | 0.9429 | 0.7928 | 0.8079 |

### 3.3 Explainable Artificial Intelligence (XAI)-Driven Analysis using SHAP

The SHAP feature importance of the best models were assessed in this study. The SHAP framework was applied on the HGB classifier with SVM-SMOTE and GRU classifier with SMOTE-ENN to estimate each predictor variable's importance in classifying bankruptcy. Larger SHAP values indicate relatively higher importance in their feature contribution. For the GRU classifier, it took a total of 1,365 iterations, a time of 2 hours, 32 minutes and 55 seconds at 6.71 seconds per iteration to generate the feature importance and the SHAP values. However, it took less than 20 seconds for the HGB model to generate the feature importance and the SHAP values.

Figure 7 shows that bankruptcy prediction in both ML and DL models was mainly influenced by a small group of financial indicators. For HGB with SVM-SMOTE, the leading predictors were Retained Earnings to Total Assets, ROA(C) before interest and depreciation before interest, Net worth/Assets, and Debt ratio %, highlighting the importance of profitability, capital structure, and solvency. For GRU with SMOTE-ENN, Debt ratio % was the strongest predictor, followed by ROA(C), Total Asset Turnover, Persistent EPS, and Operating Expense Rate, indicating that the RNN relied more heavily on leverage and operational performance in identifying bankruptcy risk.

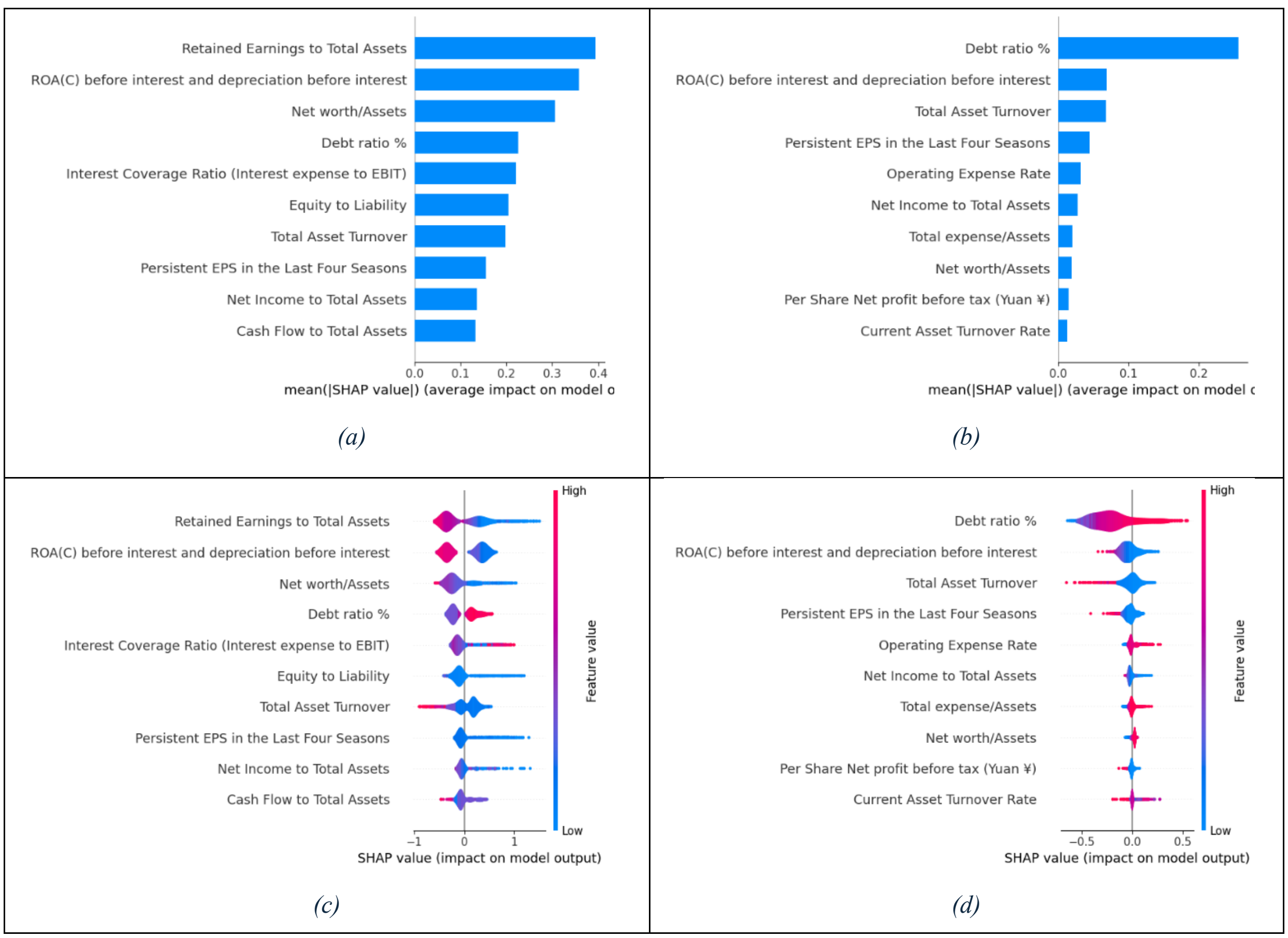

**Figure 7: Feature importance based on the mean absolute SHAP values from the HGB and GRU models. These SHAP values represent the absolute change in log odds indicating relatively higher importance with larger values. (a) Mean absolute SHAP values using HGB with SVM-SMOTE. (b) Mean absolute SHAP values using GRU with SMOTE-ENN. (c) SHAP impact model output using HGB with SVM-SMOTE. (d) SHAP impact on model output using GRU with SMOTE-ENN.**

Table 8 indicates a clear methodological progression in financial credit default and bankruptcy prediction, moving from conventional ML models such as logistic regression, decision trees, random forests, and SVMs toward more complex ensemble and deep learning architectures, including RNNs, LSTMs, and hybrid GRU-based systems. In broad terms, the more recent studies tend to report stronger discriminatory performance, particularly when evaluated with AUC, F1, PR-AUC, recall, or G-mean rather than accuracy alone. Within this pattern, the current study appears highly competitive, reporting the strongest AUC (0.9431) together with high recall (86.27%) and G-mean (84.08%), which suggests good balance between sensitivity and class-wise performance.

Table 8 also highlights an important comparability problem. The studies differ substantially in dataset type, sample size, feature dimensionality, and prediction task, ranging from credit default to bankruptcy, with instances varying from a few hundred to tens of millions.

**Table 8: Comparison of previous studies for financial credit default and bankruptcy prediction**

| Author | Year | Dataset | Features Count | Total Instances | Method | Best Results |
|---|---|---|---|---|---|---|
| [51] | 2016 | Six major USA financial institutions | 186 | 54553000 | Random forest | Average F1=81.50% |
| [52] | 2018 | German Credit Dataset | 24 | 1000 | Ensemble fusion techniques based on bstacking method | Accuracy=78.66% |
| | | Australian credit dataset | 14 | 600 | | Accuracy=88.28% |
| | | Dataset of We.com | 17 | 1421 | | Accuracy=84.07% |
| | | Lending Club dataset | 11 | 2642 | | Accuracy=66.75% |
| [53] | 2018 | Taiwan credit client's dataset | 24 | 30000 | Boosting | Accuracy=71.06% |
| [54] | 2018 | Taiwan credit client's dataset | 24 | 30000 | SPR-RIPPER | F1=53.4%, F2=88.5, AUC=70.7% |
| [55] | 2019 | Chinese consumer finance company | 490 | 44000 | Default prediction with transfer learning | AUC=0.7170, Specificity=0.6142, Sensitivity=0.7039 |
| [56] | 2019 | Taiwan credit client's dataset | 24 | 30000 | RNN-RF | AUC =0.782, Lift Index=0.659 |
| [57] | 2019 | Chinese P2P lending company dataset | 1138 | 15000 | Decision tree-based heterogeneous ensemble model | AUC=71.85% |
| [58] | 2019 | Taiwan credit client's dataset | 24 | 30000 | Ensemble model | Accuracy=83.83%, Recall=94.50%, F1=90.31% |
| [59] | 2020 | Lending Club dataset | 15 | 64139 | Logistic regression | Accuracy=79.20% |
| [60] | 2020 | Taiwan credit client's dataset | 24 | 30000 | Gradient Boosted Decision Tree Model | Accuracy=88.70% |
| | | South German credit client's dataset | 21 | 1000 | | Accuracy=83.50% |
| | | Belgium credit client's dataset | 28 | 285299 | | Accuracy=86.30% |

| [28] | 2022 | Compustat North America dataset and Center for Research in Security Prices (CRSP) dataset | 8 | 454752 | RNN, LSTM and Ensemble | AUC=0.7305 |
|---|---|---|---|---|---|---|
| [36] | 2023 | Orbis database bankruptcy dataset | 17 | 266 | FCNN | Accuracy=78.56%, AUC=0.8240 |
| [37] | 2024 | Brazil financial ratios dataset | 26 | 503 | SVM | AUC=0.7776 |
| [61] | 2025 | IND and SUM dataset | 14 | 18722059 | LSTM | Accuracy=92% |
| [62] | 2025 | Taiwan credit client's dataset | 24 | 30000 | Boruta+DBSCAN+SMOTE-Tomek+GBM | F1-score: 82.56%, G-mean: 82.98%, AUC: 0.9090, PR-AUC: 0.9185 |
| **This Study** | 2026 | Taiwan bankruptcy dataset | 95 | 6819 | Hybrid ensemble model (GRU) | G-mean: 84.08%, Recall: 86.27%, AUC: 0.9431 |

## 4 Discussions

The present study adds to the recent bankruptcy prediction literature by showing that performance in a severely imbalanced corporate-failure setting depends more on how well a model identifies the minority bankrupt class than on overall accuracy alone. First, in this study data, models such as LGBM achieved very high accuracy and specificity under SVM-SMOTE and SMOTE-Tomek, but these gains were accompanied by materially lower recall. By contrast, several deep models, especially MLP, GRU, LSTM, and RNN under SMOTE-ENN, identified a much larger share of bankrupt firms, even at the cost of more false positives. This divergence is not trivial. Recent research has shown that the ranking of bankruptcy prediction models changes when the evaluation criterion shifts from purely statistical accuracy to economically meaningful error trade-offs, because missed bankruptcies and false alarms carry asymmetric costs for creditors, investors, auditors, and regulators [63]. The current findings fit that logic closely. If the practical objective is early warning and intervention, the superior recall and G-mean of the best deep learning models may be more valuable than the higher specificity of the best boosting models.

Furthermore, hybrid resampling was not merely a preprocessing convenience but a central determinant of model behavior. SVM-SMOTE and SMOTE-Tomek tended to preserve higher specificity and accuracy, whereas SMOTE-ENN consistently produced the strongest recall and the best average G-mean across the stacked models. This pattern is consistent with the broader bankruptcy studies on imbalanced learning, which shows that aggressive editing and cleaning procedures can improve the separability of rare failure cases, even though they often do so by sacrificing some majority-class precision [7]. In this study, the practical consequence was clear: SMOTE-ENN created the conditions under which the classifiers became more responsive to financially distressed firms. This result is also in line with more recent work emphasizing that bankruptcy prediction in imbalanced settings should not rely on a single balancing recipe, because the preferred method depends on the degree of imbalance, the sample size, and the forecast horizon [64, 65]. The evidence here suggests that, for the Taiwanese dataset, the edited neighborhood structure induced by SMOTE-ENN was more compatible with the study's operational objective of minimizing missed bankruptcies.

At the level of individual model families, the results show a meaningful division of labor between ML and DL models. Among the ML models, HGB was the most consistently competitive classifier, yielding the best ROC-AUC under all three resampling strategies and comparatively strong recall relative to GB, XGB, LGBM, and AdaBoost. This suggests that histogram-based gradient boosting was particularly effective at extracting nonlinear structure from the selected ratio set while remaining relatively robust to the altered class geometry created by resampling. However, LGBM repeatedly produced the highest accuracy and specificity, indicating that it was especially conservative in assigning firms to the bankrupt class. That contrast mirrors the broader literature, where tree-ensemble methods often provide strong overall discrimination but not always the highest bankrupt-class sensitivity, especially when the training signal is sparse or heavily skewed [38, 66]. The fact that HGB rather than XGBoost emerged as the strongest ML model in this study also usefully qualifies recent evidence from other samples in which XGBoost has been dominant, suggesting that there is no universally best boosting architecture across datasets, horizons, and preprocessing pipelines [4, 38].

The DL results are equally notable. Under all three resampling strategies, the neural models generally outperformed the ML models on recall and often on G-mean, with the strongest gains emerging under SMOTE-ENN. GRU achieved the highest single-model ROC-AUC under SMOTE-ENN (0.9431), while LSTM and MLP reached the highest recall (0.8824). These results are broadly consistent with studies showing that neural architecture can be highly effective when failure risk depends on nonlinear interactions that are difficult to represent through linear or shallow tree boundaries alone [36, 39]. These findings support the usefulness of DL for bankruptcy prediction, but they also suggest that the full potential of LSTM and GRU models would probably be better assessed on multi-period firm histories rather than on one-record-per-firm data.

A particularly important result is that stacking did not automatically dominate the best standalone models. Within the stacking framework, the (GB+XGB+HGB+LGBM+AB) +LSTM coupled with SMOTE-ENN model provided the strongest overall balance, with recall of 0.7255, G-mean of 0.8254, and ROC-AUC of 0.9270. This is a credible result and confirms that heterogeneous meta-learning can still yield a practically useful compromise between sensitivity and specificity. However, several standalone deep models under SMOTE-ENN achieved substantially higher recall and higher or comparable ROC-AUC. This indicates that, in the present study, additional ensemble depth did not translate into a better early-warning frontier. One plausible explanation is limited diversity among the base learners: all five base models were boosting ensembles operating on the same tabular feature space, so the meta-learners may have received highly correlated signals. Another explanation is that the stacked architecture may have been constrained by the absence of tuned deep meta-learner hyperparameters. These are aligned with earlier evidence that stacking is most beneficial when learner diversity and feature complementarity are sufficiently strong, and that feature selection or alternative information sets can be decisive in unlocking ensemble gains [67, 68].

The feature-selection and SHAP results substantially strengthen the substantive credibility of the models. Across the consensus retention rule and the SHAP analyses, bankruptcy risk was driven primarily by retained earnings to total assets, ROA, net worth to assets, debt ratio, total asset turnover, persistent EPS, and operating expense rate. This configuration is theoretically coherent and strongly consistent with both classical bankruptcy theory and contemporary machine learning evidence. The prominence of retained earnings and ROA indicates that cumulative profitability and current earnings power remain central warning signals. The high ranking of debt ratio and net worth to assets points to the continued importance of leverage and solvency constraints, while the presence of turnover and expense indicators suggests that operational efficiency and cost discipline remain integral to the failure process. Recent SHAP-based bankruptcy studies similarly report that profitability, leverage, tax-related burdens, interest coverage, and turnover-type variables are among the most influential determinants of predicted failure risk [38, 39]. The agreement between the present SHAP results and these studies increases confidence that the models are capturing economically plausible signals rather than spurious statistical artifacts.

## 5 Conclusion & Recommendations

This study supports three broader conclusions. First, class imbalance is not a peripheral nuisance in bankruptcy prediction; it shapes model rankings, error distributions, and practical usefulness. Second, deep learning can offer substantial gains in bankrupt-class detection even in structured tabular settings, although its advantage would likely be even clearer with true sequential data. Third, explainability can meaningfully strengthen bankruptcy prediction research when it is used to validate whether influential predictors are economically interpretable and consistent across model classes. By combining consensus feature selection, hybrid resampling, heterogeneous learning architectures, and SHAP-based interpretation, the present study advances the bankruptcy studies toward a more balanced view of performance, one that values not only discrimination but also interpretability and operational relevance. That combination makes the study well aligned with the current direction of bankruptcy research, while also indicating clear next steps, especially the integration of multi-period and textual disclosures into the present framework. Accordingly, one of the most important implications of this study is not that a single algorithm universally dominates, but that different resampling-model combinations serve different institutional objectives.

A major contribution of this study is that it clarifies what can still be achieved using structured financial-ratio data alone. Recent bankruptcy research increasingly integrates textual and other non-financial inputs, including annual-report language, risk-related disclosures, sustainability reports, and multimodal corporate information, and many of these studies report incremental gains over purely numeric models [9, 10, 69, 70]. The current study does not challenge that trajectory. Rather, it shows that when feature selection, imbalance correction, ensemble learning, and XAI are carefully combined, structured accounting data remain highly informative and can still deliver strong discriminatory performance. In that sense, the present models should be seen as a robust tabular benchmark rather than as a claim that bankruptcy prediction has reached its methodological ceiling.

Indeed, recent surveys emphasize that the frontier of the field is moving toward hybrid, multimodal, and more context-aware systems, not away from financial ratios but beyond them [3, 4].

This study is not without limitations. First, the analysis was based on a single Taiwanese bankruptcy dataset, which may limit the generalizability of the results to firms in other countries, industries, or regulatory environments. Second, the models relied mainly on structured financial ratios and did not incorporate potentially useful information such as macroeconomic indicators, market-based variables, or textual disclosures from annual reports. Future studies should therefore validate the proposed framework on larger and more diverse multi-country datasets, use longitudinal and out-of-time testing designs, and integrate additional financial, market, and textual information to improve robustness, practical relevance, and generalizability.